\documentclass[letterpaper, 10 pt, journal, twoside]{IEEEtran}

\ifCLASSINFOpdf
\else
\fi

\usepackage{mypackages}
\usepackage{mydefs}

\newcommand{\pgraph}[1]{\vspace{2mm}\noindent\textbf{#1}\hspace{1mm}}

\newcommand{\rev}[1]{\textcolor{black}{#1}}

\usepackage{tikz}
\usepackage{lipsum}

\newcommand\copyrighttext{%
  \footnotesize © 2021 IEEE. Personal use of this material is permitted.  Permission from IEEE must be obtained for all other uses, in any current or future media, including reprinting/republishing this material for advertising or promotional purposes, creating new collective works, for resale or redistribution to servers or lists, or reuse of any copyrighted component of this work in other works.}
\newcommand\copyrightnotice{%
\begin{tikzpicture}[remember picture,overlay]
\node[anchor=south,yshift=2pt] at (current page.south) {\fbox{\parbox{\dimexpr\textwidth-\fboxsep-\fboxrule\relax}{\copyrighttext}}};
\end{tikzpicture}%
}

\begin{document}

\title{Weakly-Supervised Domain Adaptation \\ of Deep Regression Trackers \\ via Reinforced Knowledge Distillation}

\author{Matteo Dunnhofer, Niki Martinel, and Christian Micheloni%
\thanks{Manuscript received: December, 4th, 2020; Revised February, 25th, 2021; Accepted March, 25th, 2021.}%
\thanks{This paper was recommended for publication by Editor Eric Marchand upon evaluation of the Associate Editor and Reviewers' comments. 
This work was supported by the ACHIEVE-ITN H2020 project.} %
\thanks{All authors are with Machine Learning and Perception Lab,
        University of Udine, Udine, Italy. Matteo Dunnhofer's e-mail 
        {\tt\small matteo.dunnhofer@uniud.it}}%
\thanks{Digital Object Identifier (DOI): see top of this page.}
}

\markboth{IEEE Robotics and Automation Letters. Preprint Version. Accepted March, 2021}
{Dunnhofer \MakeLowercase{\textit{et al.}}: Domain Adaptation of Deep Regression Trackers}

\maketitle

\begin{abstract}
Deep regression trackers are among the fastest tracking algorithms available, and therefore suitable for real-time robotic applications. However, their accuracy is inadequate in many domains due to distribution shift and overfitting.
In this paper we overcome such limitations by presenting the first methodology for domain adaption of such a class of trackers. 
To reduce the labeling effort we propose a weakly-supervised adaptation strategy, in which reinforcement learning is used to express weak supervision as a scalar application-dependent and temporally-delayed feedback. At the same time, knowledge distillation is employed to guarantee learning stability and to compress and transfer knowledge from more powerful but slower trackers. 
Extensive experiments on \rev{five} different robotic vision domains demonstrate the relevance of our methodology. Real-time speed is achieved on embedded devices and on machines without GPUs, while accuracy reaches significant results.
\end{abstract}


\copyrightnotice

\IEEEpeerreviewmaketitle

\section{Introduction}
\label{sec:intro}
\begin{figure}[t]%
\centering
\includegraphics[width=\columnwidth]{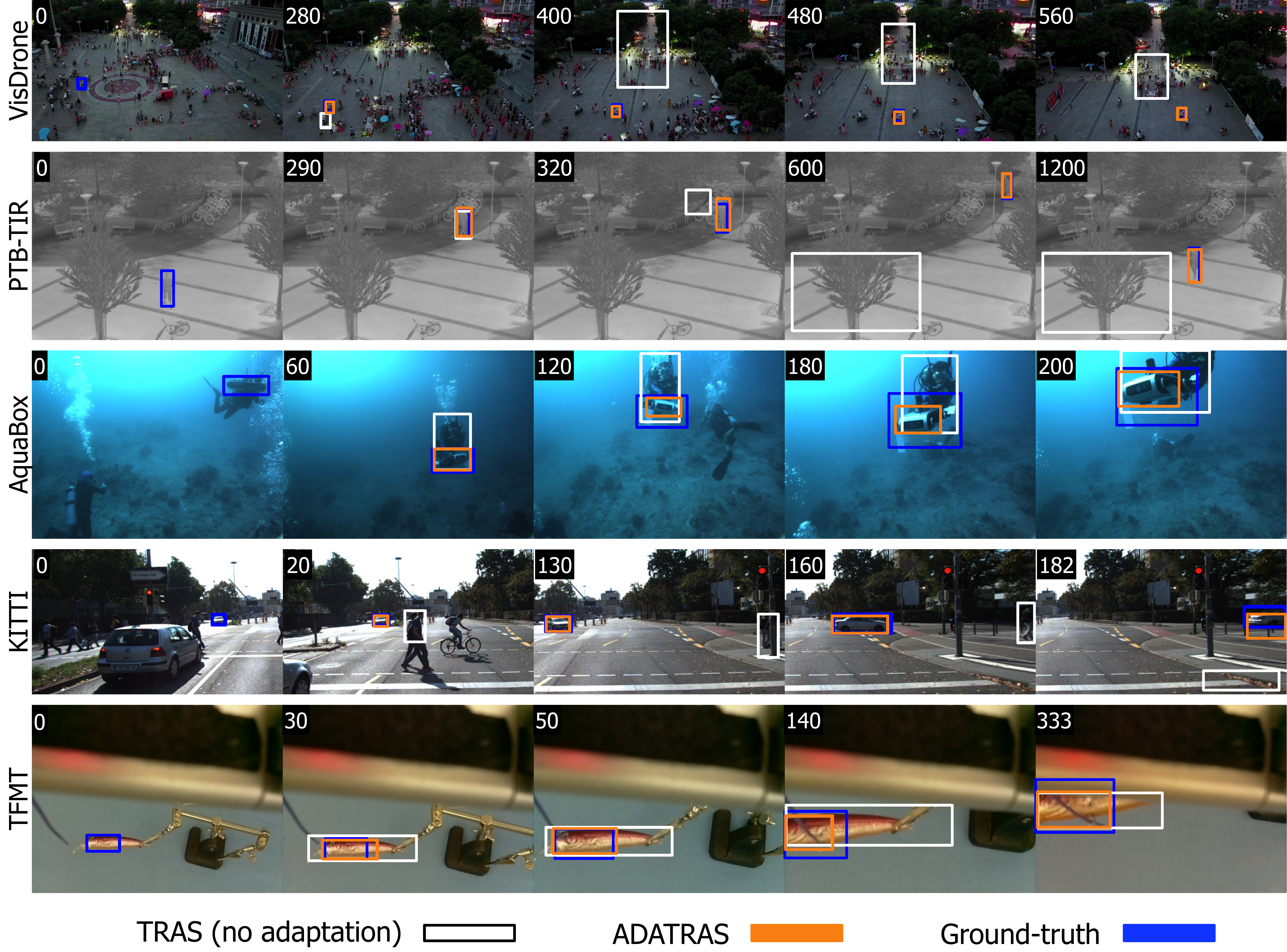}
\caption{We propose a weakly-supervised strategy to adapt DRTs for particular and small-data robotic applications. This figure shows the performance of the TRAS tracker against its adapted version ADATRAS on five different robotic tracking settings. 
For better visualization, please check out the video at this link \href{https://youtu.be/3T3BJudDSwQ}{\texttt{https://youtu.be/3T3BJudDSwQ}}.}
\label{fig:qualitative}
\end{figure}

\IEEEPARstart{R}{eal-time} visual object tracking is a key module in many robotic perception systems~\cite{Papanikolopoulos1991,Portmann2014,KITTI,AquaBox,Luiten2019ral,Dunnhofer2020MedIA}.
Recently, deep regression trackers \cite{RE3,GOTURN,Dunnhofer2020accv} (DRTs) have been proposed in the robotics community \cite{RE3} because of their efficiency and generality. Thanks to their simple architecture, DRTs achieve processing speeds that surpass 100 FPS, making them suitable even for low-resource robots. Moreover, with the availability of large-scale computer vision datasets \cite{ImageNet}, these trackers can learn to track a large variety of targets without relying on particular assumptions, thus simplifying the development of tracking pipelines.
However, acquiring thousands of videos for training these systems is not realistic in many real-world robotic application domains. 
Additionally, many domains offer particular scenarios that differ much from the examples which DRTs are trained on. For example, drone \cite{Chaudhary2017} and driving \cite{KITTI,RoadText1k} applications require tracking objects from particular camera views. Underwater robots offer uncommon targets and settings \cite{AquaBox,Langis2020}. Other robotics systems can use different imaging modalities \cite{Portmann2014}. Robotic manipulation configurations need the tracking of atypical objects \cite{TFMT}.
As shown in Figure~\ref{fig:qualitative}, these situations cause DRTs' accuracy to be very low. This is due to their deep learning architecture that is subject to overfitting if trained directly on small application datasets, and suffers from the shift between training and test data distributions when trained for large-scale generic object tracking.

To address these issues, the visual tracking community proposes to increase the learning capacity of convolutional neural networks \cite{SiamRPNpp} (CNNs), or to use online learning mechanisms to adapt the capabilities of deep trackers~\cite{MDNet,ATOM,DiMP} to every new target in every new video. These strategies lead to higher accuracy and robustness, but at cost of real-time speed achieved just on high-end machines.
On the other hand, transfer learning \cite{Pan2010} and domain adaptation \cite{Csurka2017} are widely used machine learning techniques to address such issues.
The idea is to exploit the knowledge acquired in a source domain and adapt it to new target domains through an additional offline learning stage that exploits a few examples of the target domain. This allows to improve performance and generalization on the new domains, without sacrificing the test-time processing speed, as the deep models can be applied without any additional tuning.
Different solutions have been proposed to adapt robotic vision systems \cite{Angeletti2018,Zhang2019ral,Carlson2019}, but no work considered adaptation in the context of robotic tracking.

Considering these motivations, 
the main contribution of this paper is the first methodology for offline domain adaption \rev{of DRTs.
This is also the first work that considers the domain adaptation problem in the context of visual tracking.}
To reduce the labeling effort and maintain application-specific development, we propose a weakly-supervised adaptation procedure. 
Thanks to reinforcement learning (RL), the knowledge previously acquired in a generic object tracking domain is adjusted with scalar signals that can be also delayed in time. 
But, as RL optimization is difficult, we build upon the experience of more accurate but slower trackers via knowledge distillation (KD) to stabilize learning and additionally improve the performance.
We build our solution on the recent framework proposed in~\cite{Dunnhofer2020accv}, which marries KD and RL for generic object tracking. However, such a method is designed for learning DRTs with bounding-box level and densely annotated datasets. Hence, as an additional contribution, in this paper we offer a generalization of~\cite{Dunnhofer2020accv} that allows its exploitation in weakly labeled settings and for generic application objectives.
\rev{Extensive analysis on five different robotic tracking domains shows that the proposed adaptation strategy makes DRTs applicable again on particular and low-resource robotic perception domains.}

\vspace{-0.02cm}
\section{Related Work}

\pgraph{Domain Adaptation in Robotic Vision.}
\rev{Domain adaptation has been previously studied in robotic vision. 
Spatial information about the domains has been exploited to adapt robotic vision system to recognize new objects \cite{Angeletti2018}.
Wulfmeier et al. \cite{Wulfmeier2017} used adversarial approaches to adapt segmentation models to the visual appearance of weather and seasonal conditions.
Batch normalization layers have been exploited for robotic vision-based kitting \cite{Mancini2018iros}.
Particular loss functions \cite{Fang2018,Zhang2019ral}, augmentation networks \cite{Carlson2019}, or pretext tasks \cite{Loghmani2020} have been proposed extensively for the adaptation of visual capabilities from simulated to real environments.
Bellocchio et al. \cite{Bellocchio2020} used generative adversarial networks to adapt robotic fruit counting systems to unseen species.
Yet, no work considered the problem of adapting tracking knowledge acquired in a generic domain to another different robotic target domain. Furthermore, no method mixing RL and KD has been introduced in the context of domain adaptation.
}

\pgraph{Adaptation in Visual Tracking.}
In the visual tracking panorama, the concept of domain adaptation has been used to refer to instance-level online learning performed on the target object of every new test sequence.
MDNet-based trackers~\cite{MDNet} consider a training sequence as a domain and propose a CNN learned offline via binary classification on multiple domain-specific branches. Such a model is then refined on every test sequence by solving an online binary classification problem.
ATOM~\cite{ATOM} combines an offline learned bounding-box regression model with an efficient target-background IoU-based discriminator which is trained exclusively online.
DiMP~\cite{DiMP} performs an online update of a target-specific CNN model via a fast discriminative-learning optimization strategy.
With respect to these works, our solution is conceptually different. We consider the target domain as a set of videos whose a subset is dedicated to offline learning. This introduces an additional training procedure, but it allows the adapted tracker to be extremely fast at test time, thanks to the avoidance of online adaptation mechanisms that reduce the tracking speed.

\section{Methodology}

We build our solution upon the recent framework of \cite{Dunnhofer2020accv}, which showed the effectiveness of combining KD and RL for generic object tracking. 
Differently from \cite{Dunnhofer2020accv}, in this paper, we use RL to express a weak and temporally-delayed application-specific objective and employ KD to make the convergence achievable. 
We remark that our the methodology of \cite{Dunnhofer2020accv} is not applicable as it is for our problem, because it assumes a fully supervised setting in which dense ground-truth bounding-box data is available.

\begin{figure}[t]%
\centering
\includegraphics[width=.75\columnwidth]{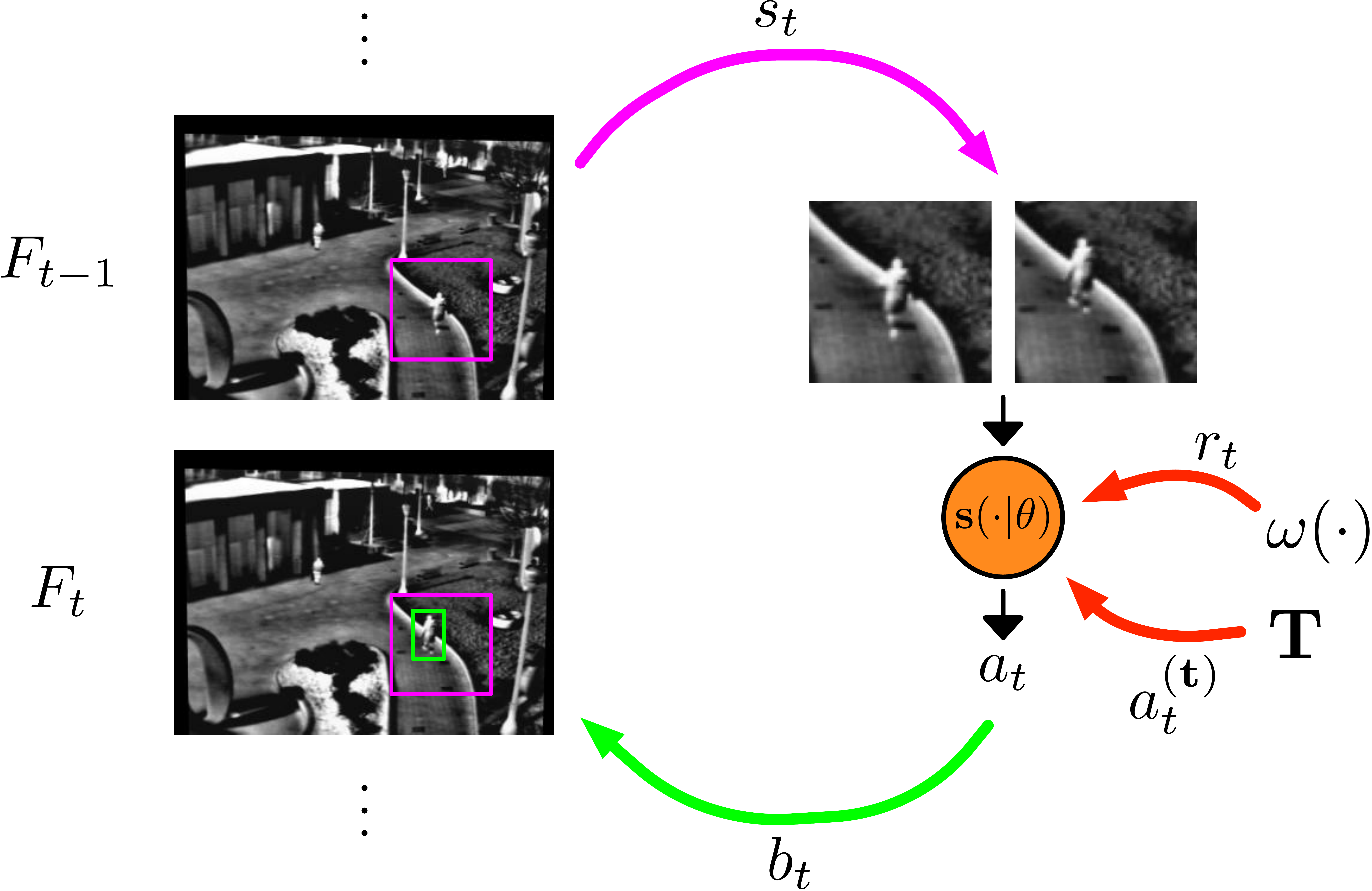}
\caption{Visual representation of the MDP interaction process between the student and a video. At each step $t$, a state $\state_t$ is extracted from frames $\frame_{t-1}, \frame_{t}$. $\state_t$ is processed by $\student$ which outputs the action $\action_t$ that is transformed into the bounding-box output $\bbox_t$. During the adaptation procedure, the learning is driven by the weak supervision function $\weaksup(\cdot)$ and by the actions of the set of teacher trackers $\teachers$.}
\label{fig:processing}
\end{figure}

\pgraph{Preliminaries and Problem Statement.}
We consider a video $\video_j = \big\{ \frame_t \in \images \big\}_{t=0}^{T_j}$ as a sequence of frames, where each $\frame_t$ belongs to the space of RGB images $\images = \{0,\cdots,255\}^{w \times h \times 3}$. Each video has a target object to be tracked, which is defined in the first frame $\frame_0$ through a bounding-box $\bboxgt_0 = [x_0^{(g)},y_0^{(g)},w_0^{(g)},h_0^{(g)}] \in \reals^4$ that specifies the coordinates of the object's top left corner, and its width and height. The goal of a tracker, given each frame $\frame_t$, is to predict the bounding-box $\bbox_t = [x_t,y_t,w_t,h_t]$ that best fits the target in $\frame_{t}$.

As our solution is based on the KD framework, we employ the concepts of student and teacher \cite{Hinton2014KD}.
We formally consider a regression-based tracker 
as the student 
$\student : \images \times \images \times \Theta \rightarrow \reals^4$
which is a function parameterized by $\weights \in \Theta$ that outputs the relative motion of the target contained in two consecutive images given as input. 
We assume the student has acquired general tracking knowledge by optimizing $\weights$ on the videos of a source domain $\sourceset$.
The set of teachers is defined as
$\teachers = \big\{ \teacher : \images \rightarrow \reals^4 \big\}$
where each $\teacher$ is a generic tracking function that, given a frame, produces a bounding-box for that frame.

Our problem of interest consists in adapting $\student(\cdot|\weights)$'s past ability to a new tracking domain $\targetset$ different from $\sourceset$. More specifically, we assume $\targetset$ is split into a training set $\targetsettrain \subseteq \targetset$, and a test set  $\targetsettest \subseteq \targetset$ with $\targetsettrain \cap \targetsettest = \emptyset$. The goal is to exploit $\targetsettrain$, for which weak supervision is given, to maximize the tracking performance on the videos of $\targetsettest$.

\pgraph{Video Processing.}
\label{sec:votmdp}
To use RL, we model the tracking as an interaction process~\cite{SuttonBarto2018}. We treat $\student(\cdot|\weights)$ as an artificial agent which interacts with a video $\video_j \in \targetset$ according to the Markov Decision Process (MDP) definition given in~\cite{Dunnhofer2020accv}.
The interaction happens as a temporal sequence of states $\state_1, \state_2, \cdots, \state_t$, and actions $\action_1, \action_2, \cdots, \action_t$.
Every $\state_t$ is defined as a pair of image patches obtained by cropping $\frame_{t-1}$ and $\frame_t$ using the previously known bounding-box $\bbox_{t-1}$ and a factor $\contextfactor$ that enlarges the patches in order to include additional image context information.
At the $t$-th step, the student is given the state $\state_t$ and outputs the continuous action $\action_t$.
Each $\action_t$ 
is defined as the vector $\action_t = [\Delta x_t, \Delta y_t, \allowbreak \Delta w_t, \Delta h_t] \in [-1,1]^4$ which quantifies the relative horizontal and vertical translations ($\Delta x_t, \Delta y_t$, respectively) and width and height scale variations ($\Delta w_t, \Delta h_t$, respectively) that have to be applied to $\bbox_{t-1}$ to predict $\bbox_t$.
Hence, based on the previous bounding-box estimate \cite{Dunnhofer2020accv}, $\action_t$ is transformed into the bounding-box $\bbox_t$ which provides the localization of the target in frame $\frame_t$. 
A visual representation of the interaction procedure is depicted in Figure \ref{fig:processing}.

\begin{figure}[t]%
\centering
\includegraphics[width=\columnwidth]{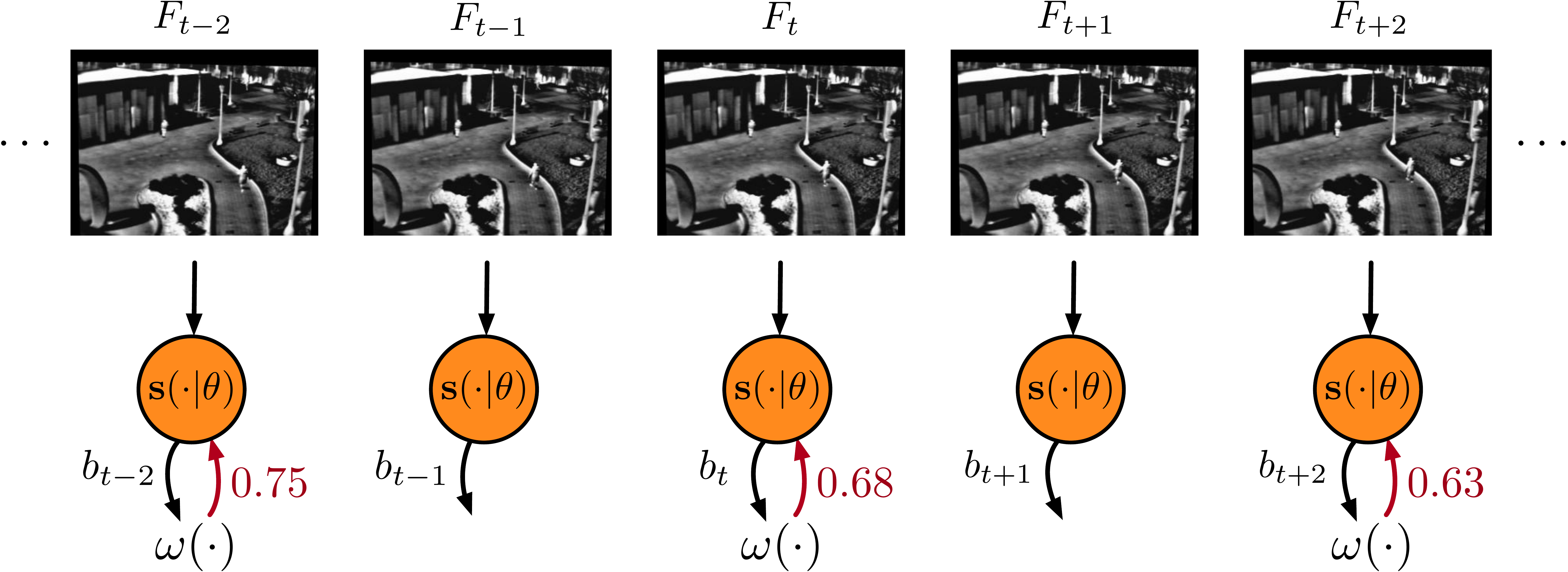}
\caption{Visual representation of the student's feedback mechanism based on the weak supervision function $\weaksup(\cdot)$. At every $t$, $\student(\cdot | \weights)$ gives its bounding-box prediction $\bbox_t$ which is eventually evaluated with a 0-1 score by $\weaksup(\cdot)$ (if $\weaksup(\cdot)$ is defined for that temporal step).}
\label{fig:weaksuptempvisual}
\end{figure}

\pgraph{Weak Supervision.} 
During adaptation on videos $\video_j \in \targetsettrain$, the actions $\action_t$ of $\student$ are rewarded by the scalar value $\reward_t \in [-1, 1]$ \rev{(in RL terms, the reward)}. 
In our setting, this is what we use to express weak supervision. 
Differently from \cite{Dunnhofer2020accv}, who proposed a continuously available bounding-box overlap formulation, we just assume the feedback to be released as a 0-1 value through an arbitrary function $\weaksup : \reals^4 \rightarrow [0, 1]$ that evaluates a bounding-box prediction $\bbox_t$ and that can be implemented based on the application needs. Additionally, we do not require $\weaksup(\bbox_t)$ to be defined for every $t$.
$\weaksup(\bbox_t)$ is formally exploited in our proposed \rev{MDP} reward definition which, at every $t$, is
\begin{align}
\reward_t = \reward(\bbox_t) = 
    \begin{cases}
    0 \ \text{if } \weaksup(\bbox_t) \text{ is not defined } \\
    \nu\left(\weaksup(\bbox_t)\right) \ \text{if } \weaksup(\bbox_t) \text{ is defined } \wedge \weaksup(\bbox_t) \geq 0.5 \\
    -1 \ \text{otherwise}
    \end{cases}
\end{align}
with
$\nu(z) = 2(\lfloor z \rfloor_{0.05}) - 1$ that
floors to the closest $0.05$ digit and shifts the input range from $[0,1]$ to $[-1,1]$. 
Figure \ref{fig:weaksuptempvisual} visualizes the proposed weak supervision mechanism.

\begin{table*}[t]
\fontsize{7}{8}\selectfont
	\centering
	\caption{\rev{Statistics of the target domains selected for this work. The number of training and test videos, frames, and the number of sequences after splitting the videos in chunks of 32 frames, are reported in the first five rows. $\ss$ and $\ps$ obtained by the teachers on the training videos are reported in the last three rows.} }
	\label{tab:domainsstats}
	\setlength\tabcolsep{.15cm}
	\begin{tabular}{l | c c | c c | c c | c c | c c}
		\toprule
		Target Domain & \multicolumn{2}{c|}{\visdrone} & \multicolumn{2}{c|}{\ptbtir} & \multicolumn{2}{c|}{\aquabox} & \multicolumn{2}{c|}{KITTI} & \multicolumn{2}{c}{TFMT}\\
		
		\midrule
		
		$\big|\targetsettrain\big|$ & \multicolumn{2}{c|}{86} & \multicolumn{2}{c|}{48}  & \multicolumn{2}{c|}{41} & \multicolumn{2}{c|}{21} & \multicolumn{2}{c}{6} \\
		$\big|\targetsettest\big|$ & \multicolumn{2}{c|}{11} & \multicolumn{2}{c|}{12}  & \multicolumn{2}{c|}{20}  & \multicolumn{2}{c|}{20} &  \multicolumn{2}{c}{6} \\
		\# frames $\targetsettrain$ & \multicolumn{2}{c|}{69941} & \multicolumn{2}{c|}{23497}  & \multicolumn{2}{c|}{3927}  & \multicolumn{2}{c|}{4797}  & \multicolumn{2}{c}{520} \\
		\# frames $\targetsettest$ & \multicolumn{2}{c|}{7046} & \multicolumn{2}{c|}{6532}  & \multicolumn{2}{c|}{6033}  & \multicolumn{2}{c|}{4143} & \multicolumn{2}{c}{1320} \\
		\# splitted sequences $\targetsettrain$ & \multicolumn{2}{c|}{1696} & \multicolumn{2}{c|}{804}  & \multicolumn{2}{c|}{263} & \multicolumn{2}{c|}{356} & \multicolumn{2}{c}{42}  \\
		
		\midrule
		$\tm$ $\targetsettrain$ $\ss$ $\ps$ & 0.556 & 0.798 & 0.565 & 0.817 & 0.321 & 0.488 & 0.385 & 0.668 & 0.283 & 0.385 \\
		$\ts$ $\targetsettrain$ $\ss$ $\ps$ & 0.576 & 0.741 & 0.617 & 0.785 & 0.499 & 0.717 & 0.430 & 0.579 & 0.460 & 0.659 \\
		$\ta$ $\targetsettrain$ $\ss$ $\ps$ & 0.555 & 0.759 & 0.559 & 0.691 & 0.563 & 0.843 & 0.450 & 0.619 & 0.619 & 0.783 \\
		
		\bottomrule		
\end{tabular}
\end{table*}

\pgraph{Adapting the Tracker.}
The student's parameters $\weights$, which have been pretrained on the $\sourceset$, are adapted to $\targetset$ by learning offline on $\targetsettrain$. 
To do this, we employ the end-to-end strategy proposed in \cite{Dunnhofer2020accv} and we briefly report it by highlighting the improvements that allow its generalization for weak supervision.

Our adaptation strategy provides two learning objectives that are fulfilled at the same time. To optimize the actions with respect to $\weaksup(\cdot)$ (by means of the rewards), the following RL actor-critic loss formulation \cite{Sutton2000}
\begin{gather}
\label{eq:rlloss}
    \lossrl = \policyloss + \valueloss \\
\label{eq:policyloss}
\policyloss = -\sum_{i=1}^{t_{max}} \log \student(\state_i | \theta)\big(r_i + \gamma \student_{\statevalue}(\state_{i+1} | \theta) - \student_{\statevalue}(\state_i | \theta) \big) \\
\label{eq:valueloss}
\valueloss = \sum_{i=1}^{t_{max}} \frac{1}{2} \big(R_i -  \student_{\statevalue}(\state_i | \theta) \big)^2, 
R_i = \sum_{k=1}^{i}\gamma^{k-1}\reward_k
\end{gather}
is applied after $t_{max}$ steps of interaction with $\video_j \in \targetsettrain$, in which each $\action_t$ performed by $\student$ is sampled from a normal distribution $\mathcal{N}(\mu, \sigma)$. To attend this optimization goal, the student is set to produce the additional output $\statevalue_t = \student_{\statevalue}(\state_t| \weights)$, which is the prediction of the \rev{$\gamma$-discounted} cumulative reward \rev{$R_i$} that $\student$ expects to receive from $\state_t$ to the end of the interaction. 
\rev{In RL terms, $\policyloss$ and $\valueloss$ are known as policy gradient loss with advantage and value loss respectively.}

On a second side, our adaptation scheme minimizes the following objective 
\begin{align}
\label{eq:distloss}
\distillationloss &= \sum_{i=1}^{t_{max}} |a^{(\teacher)}_i - \student(\state_i | \weights)| \cdot m_i, 
\end{align}
which is the L1 loss \cite{RE3,GOTURN} between the actions performed by $\student$ and the actions $a^{(\teacher)}_t$ that the teacher would take to move $\student$'s bounding-box $\bbox_{t-1}$ into the $\teacher$'s prediction $\bbox^{(\teacher)}_t$ \cite{Dunnhofer2020accv}. 
\rev{Each of the differences in Eq. (\ref{eq:distloss}) are multiplied by the binary values $m_i$ which represent the case in which $\student$ performs worse than the teacher.}
This learning objective makes the learning feasible and has the additional advantage of extracting knowledge from more accurate and robust tracking algorithms, leading ultimately to better performance.
A distributed setting \cite{Gorila}
is employed to implement the overall optimization strategy by considering $\frac{S}{2}$ students for the optimization of Eq. (\ref{eq:rlloss}) and the other $\frac{S}{2}$ for Eq. (\ref{eq:distloss}).

The proposed adaptation procedure brings some modifications to the learning method of \cite{Dunnhofer2020accv} that it allows to work in weakly supervised settings.
First, the \rev{policy gradient} $\log \student(\state_i | \theta)$ term used in Eq. (\ref{eq:policyloss}) is obtained after the definition of a normal distribution $\mathcal{N}(\mu, \sigma)$ with mean defined as $\mu = \student(\state_t | \theta)$ and standard deviation $\sigma$ considered with a fixed value. In this way, varying $\sigma$ one can control $\student$'s exploration without the need of ground-truth bounding-boxes as in \cite{Dunnhofer2020accv}. 
Second, we propose to favor $\teacher$'s prediction in the computation of $m_i \in \{0,1\}$ of Eq. (\ref{eq:distloss}), by setting $m_i = 1$ if $\reward(\bbox_t^{(\teacher)}) \geq \reward(\bbox_t)$ holds and $m_i = 0$ otherwise. This in order to address the 0 reward scenarios caused by the non definition of $\weaksup(\cdot)$, in which it is not possible to infer the actual student performance.
Third, the $\teacher$ to which learn from in Eq. (\ref{eq:distloss}) is selected before the start of the interaction via some arbitrary decision strategy that can be defined depending on the application objectives and resources.

\pgraph{Tracking after Adaptation.}
After the adaptation-by-learning process is done, the student $\student(\cdot | \weights)$ is ready to be used for tracking on $\targetsettest$ as follows. 
We consider each testing video $\video_j \in \targetsettest$, for which the target is individuated in $\frame_0$ by the bounding-box $\bboxgt_0$, as the aforementioned MDP. At each $t$, $\state_t$ are extracted from $\frame_{t-1}, \frame_t$, and $\action_t$ are performed by means of the student's adapted policy $\student(\state_t | \weights)$ and  transformed into bounding-box outputs $\bbox_t$. 
We name the tracker resulting from this tracking procedure ADATRAS (ADApted TRAcking Student).

\section{Experimental Setup}

\pgraph{Performance Measures.}
To evaluate the performance of the trackers involved in this paper, we follow the standard methodology introduced in \cite{OTB}.
Trackers are initialized in the first frame of a sequence with the target ground-truth bounding-box and let run until the end, respecting the one-pass evaluation (OPE) protocol. The quantitative measures used are the area under the curve (AUC) of the success and precision plots, which are referred as to success score ($\ss$) and precision score ($\ps$) respectively.

\pgraph{Tracker.}
We follow the most recent advancements in deep regression tracking \cite{RE3,Dunnhofer2019,Dunnhofer2020accv} to implement our tracker $\student(\cdot | \weights)$ as a deep neural network with weights $\weights$.
The network gets as input $\state_t$ as two image patches which pass through two ResNet-18 \cite{He2016ResNet} CNN branches with shared weights. The subsequent feature maps are linearized, concatenated together, and fed to two consecutive fully connected layers with ReLU activations and an LSTM layer \cite{Hochreiter1997LSTM}, both with 512 neurons. The LSTM's output is finally fed to two fully connected heads that output the action $\action_t = \student(\state_t | \theta)$ and the state-value $\statevalue_t = \student_{\statevalue}(\state_t | \theta)$ respectively.

\pgraph{\rev{Source and Target Domains.}}
We conducted experiments considering the GOT-10k \cite{GOT10k} and LaSOT \cite{LaSOT} benchmarks as source domains $\sourceset$. These are large-scale tracking datasets containing, respectively, 10000 and 1400 videos of generic target objects (up to 563 to different object classes) in generic tracking settings.
The initial optimization of $\weights$ on these sets was performed following the details of \cite{Dunnhofer2020accv}.

\begin{table*}[t]
\fontsize{7}{8}\selectfont
	\centering
	\caption{\rev{Comparison between ADATRAS and the teachers and DRTs. FPS are obtained on the \server\ machine. Best results, per domain and performance measure, are highlighted in red, second-best in blue, third-best in green.}}
	\label{tab:accuracy}
	\begin{tabular}{l | c c c | c c c | c c c | c c c | c c c}
		\toprule
		\multirow{2}{*}{Tracker} & \multicolumn{3}{c|}{\visdrone} & \multicolumn{3}{c|}{\ptbtir} & \multicolumn{3}{c|}{\aquabox} & \multicolumn{3}{c|}{KITTI} & \multicolumn{3}{c}{TFMT}\\
                    & $\ss$ & $\ps$ & FPS & $\ss$ & $\ps$ & FPS & $\ss$ & $\ps$ & FPS & $\ss$ & $\ps$ & FPS & $\ss$ & $\ps$ & FPS \\
		\midrule
		MDNet \cite{MDNet} & \tblbest{0.559} & \tblbest{0.902} & 2 & 0.586 & \tblbest{0.953} & 2 & 0.543 & 0.504 & 2 & 0.413 & \tblsecondbest{0.686} & 3 & \tblsecondbest{0.501} & \tblbest{0.770} & 2  \\
		SiamRPN++ \cite{SiamRPNpp} & 0.532 & 0.790 & 31 & \tblthirdbest{0.614} & 0.774 & 55 & \tblsecondbest{0.591} & \tblbest{0.753} & 41 & \tblthirdbest{0.504} & \tblthirdbest{0.632} & 43 & 0.504 & 0.637 & 27  \\
		ATOM  \cite{ATOM} & \tblthirdbest{0.539} & \tblsecondbest{0.891} & 16 & \tblsecondbest{0.620} & \tblthirdbest{0.778} & 24 & \tblbest{0.594} & \tblsecondbest{0.742} & 20 & \tblsecondbest{0.529} & \tblsecondbest{0.686} & 19 & \tblbest{0.615} & \tblthirdbest{0.639} & 28 \\
		
		\midrule
		
		GOTURN  \cite{GOTURN} & 0.350 & 0.600 & 56 & 0.327 & 0.497 & \tblsecondbest{200} & 0.402 & 0.468 & \tblthirdbest{110} & 0.209 & 0.313 & 62 & 0.363 & 0.271 & \tblsecondbest{125}  \\
		RE3  \cite{RE3} & 0.354 & 0.626 & \tblthirdbest{60} & 0.201 & 0.278 & \tblbest{255} & 0.445 & 0.398 & 107 & 0.221 & 0.308 & \tblthirdbest{63} & 0.406 & 0.272 & \tblbest{136}  \\
		TRAS \cite{Dunnhofer2020accv} & 0.384 & 0.653 & \tblsecondbest{65} & 0.432 & 0.603 & 168 & 0.522 & 0.619 & \tblsecondbest{161} & 0.486 & 0.627 & \tblsecondbest{85} & 0.332 & 0.169 & 112 \\
		\midrule
		
		ADATRAS & \tblsecondbest{0.552} & \tblthirdbest{0.823} & \tblbest{67} & \tblbest{0.661} & \tblsecondbest{0.862} & \tblthirdbest{170} & \tblthirdbest{0.576} & \tblthirdbest{0.732} & \tblbest{165} & \tblbest{0.537} & \tblbest{0.720} & \tblbest{89} & \tblthirdbest{0.581} & \tblsecondbest{0.659} & \tblthirdbest{115} \\
		\bottomrule		
\end{tabular}
\end{table*}

We demonstrate the capabilities of our solution on \rev{five robotic} target domains, which we refer to as \visdrone, \ptbtir, \aquabox , \rev{KITTI, and TFMT.}
These were selected due to their particular characteristics in: camera views; uncommon objects and motions; image modality. Statistics of the domains are shown in Table \ref{tab:domainsstats}. Beside being real-world \rev{robotic vision} datasets, these domains also contain bounding-box labels for every frame of the videos. This 
permits an accurate validation with the control and simulation of the loss of supervision.

\paragraph{\visdrone} This domain concerns tracking objects in videos acquired from drones, and it is based on the publicly available VisDrone 2019 challenge dataset \cite{VisDrone2019}. Targets available are persons, cars, or animals, acquired by particular camera views in which they appear very small and their motion is different depending on the drone's altitude. To implement $\targetsettrain$ and $\targetsettest$ we employed, respectively, the original training and validation sets provided by the authors \cite{VisDrone2019}.

\paragraph{\ptbtir} The \ptbtir\ target domain regards tracking people in videos acquired through a thermal-infrared (TIR) camera. This domain offers common objects (people), but video frames are represented via a different sensor. The data contained in the PTB-TIR benchmark \cite{PTBTIR} was employed. $\targetsettrain$ and $\targetsettest$ were obtained by randomly splitting, with an 80-20 ratio, the 60 videos contained in the benchmark.

\paragraph{\aquabox} This domain consists in tracking an underwater robot in an underwater video setting. Videos offer targets, camera views, motions, and object physics, that are very unusual from what available in standard tracking datasets. 
The data used was obtained with the AquaBox dataset \cite{AquaBox}.
For $\targetsettrain$ and $\targetsettest$ we employed the training and validation sets provided by the authors. Only videos composed of at least 25 frames were retained.

\paragraph{\rev{KITTI}} 
\rev{This domain concerns tracking vehicles and people in videos acquired from a vehicle point of view, thus offering new camera views (different from the drone's) on common objects. We used the popular KITTI dataset \cite{KITTI} to implement $\targetsettrain$ and $\targetsettest$. We considered tracks longer than 100 frames of the KITTI's training videos and splitted them with a 50\% ratio. }

\paragraph{\rev{TFMT}}
\rev{This target domain requires tracking objects to perform a fine grained manipulation task with a robotic arm \cite{TFMT}. The targets and the settings contained in this dataset are very uncommon to generic object trackers. Tthe 12 labeled videos contained in the TFMT dataset \cite{TFMT} have been splitted with a 50\% ratio to implement $\targetsettrain$ and $\targetsettest$.}

\begin{figure}[t]%
\centering
\includegraphics[width=\columnwidth]{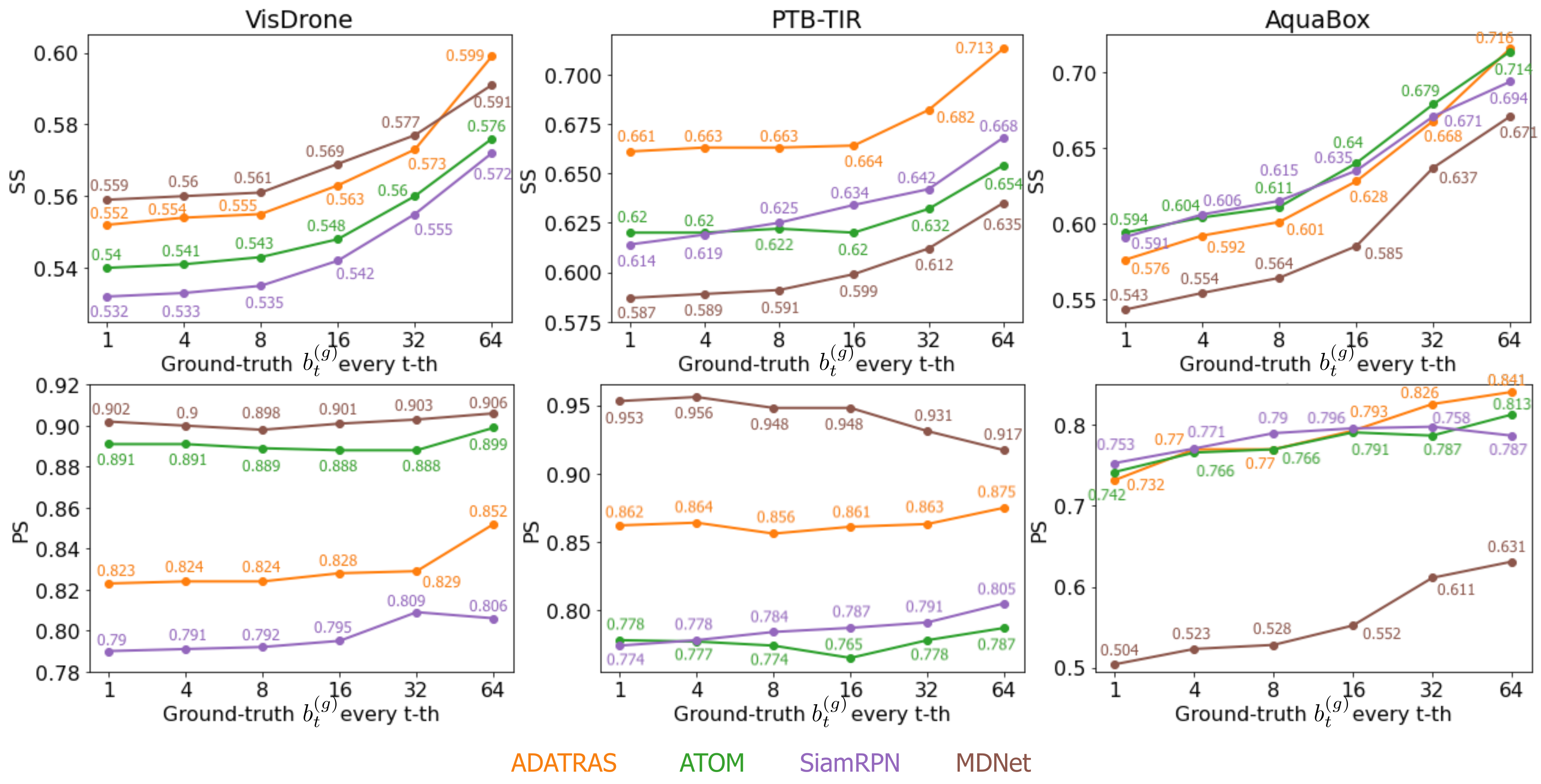}
\caption{Effect of evaluating the trackers with a weakly-labeled test set in which ground-truths are available every 4-th, 8-th, 16-th, 32-th, and 64-th frame.}
\label{fig:evaltemp}
\end{figure}

\pgraph{Weak Supervision Form.}
We experimented two forms of 0-1 function to weakly supervise  $\student(\cdot|\weights)$. The first implements $\weaksup$ as the function $\weaksupiou(\bbox_t, \bboxgt_t) = \text{IoU}(\bbox_t, \bboxgt_t)$, which takes the student's predicted bounding-box and a bounding-box reference $\bboxgt_t$ of the target and computes their intersection-over-union as 
\begin{align}
\text{IoU}(\bbox_t, \bboxgt_t) = (\bbox_t \cap \bboxgt_t) / (\bbox_t \cup \bboxgt_t). %
\end{align}
The second form is through the 0-1 function $\weaksupdist(\bbox_t, \bboxgt_t) = 1 - \text{NormDist}(\bbox_t, \bboxgt_t)$ where 
\begin{align}
\text{NormDist}(\bbox_t, \bboxgt_t) = \frac{\left\lceil{\sqrt{(x_t - x_t^{(g)})^2 + (y_t - y_t^{(g)})^2}}\right\rceil_{20}}{20}
\end{align}
is the function that computes the pixel distance between the centers of $\bbox_t$ and $\bboxgt_t$, truncated at 20 and normalized by the same value. The value 20 was chosen following the standard precision score threshold defined in \cite{OTB}.

\pgraph{Teachers.}
The tracking teachers selected for this work are MDNet \cite{MDNet}, SiamRPN++ \cite{SiamRPNpp}, and ATOM \cite{ATOM}.
Since they tackle visual tracking by different approaches, it is more likely that at least one can succeed in the application domain, and thus provide useful tracking knowledge.
In the experiments, we considered exploiting single teachers or a pool of teachers. In particular, the following sets of teachers were examined $\tm = \{ \tmdnet \}, \ts = \{ \tsiamrpn \}, \ta = \{ \tatom \}, \tp = \{ \tmdnet, \tsiamrpn, \tatom \}$. The performance of these on $\targetsettrain$ are shown in the last three rows of Table \ref{tab:domainsstats}.
We studied two methods to select the teacher $\teacher$ in Eq. (\ref{eq:distloss}). The first randomly selects $\teacher \in \teachers$ using a uniform distribution. The second selects $\teacher$ based on the average $\weaksup(\cdot)$ performance, that is
\begin{align}
\label{eq:teacherselection}
\teacher \in \teachers \: : \: \Omega(\teacher, \video_j) = \max_{\teacher \in \teachers} \Omega(\teacher, \video_j)
\end{align}
where 
\begin{align}
\Omega(\teacher, \video_j) = \frac{\sum_{t=0}^{T_j}\weaksup(\bbox_t^{(\teacher)})}{T_j}
\end{align}
is a 0-1 number that estimates the quality of $\teacher$'s predictions given for video $\video_j \in \targetsettrain$.

\pgraph{Implementation Details.}
To produce more training samples, each video (and the respective filtered bounding-box sequences) was split in 20 randomly indexed sequences of 32 frames, following \cite{RE3,Dunnhofer2020accv}. The total number of videos is reported in row five of Table \ref{tab:domainsstats}. A temporal reverse of the sequences was also applied with 50\% probability during training.
$S = 12$ training students were used for training. 
$\sigma = 0.05$ was set for the \visdrone\, \ptbtir, KITTI, and TFMT domains, and $\sigma = 0.025$ for \aquabox.
To facilitate the learning, a curriculum learning procedure similar to \cite{RE3} was employed by increasing the length of the interaction during the adaptation procedure. 
The Adam optimizer~\cite{Kingma2014} was utilized. A learning rate of $7.5\cdot 10^{-7}$ was set for all the layers of the student except for the fully connected layer that predicts $\statevalue_t$, for which learning rate was set to $10^{-5}$. 
The student was trained until the validation performance stopped improving.
Trainings took between 12 and 48 hours, depending on the amount of data available in a domain.
We experienced some variance between the outcomes of different experiment runs, and therefore we report averaged results.
Other settings not specified in this section have been inherited from \cite{Dunnhofer2020accv}.
In the following of the paper, if not specified otherwise, default experimental settings are with the student initially optimized on GOT-10k-based $\sourceset$, learning from $\tp$ by respecting Eq. (\ref{eq:teacherselection}), with weak supervision based on $\weaksupiou(\cdot)$ given at every time step, and with the hyper-parameters mentioned above.

\pgraph{Hardware and Software.}
We employed three hardware machines for our experiments. A high-end server machine with an Intel Xeon E5-2690 v4 @ 2.60GHz CPU, 320 GB of RAM, and 4 NVIDIA TITAN V GPUs. We refer to this as \server. A desktop computer with an Intel Xeon W-2125 @ 4.00GHz CPU, 32GB of RAM, and an NVIDIA GTX1080-Ti, which we refer to as \desktop. And an embedded board NVIDIA Jetson Nano with an ARM A57 quad-core 1.43 GHz CPU, 4GB of RAM, and a Maxwell 128 core GPU, which we refer to as \embedded. 
All the code was implemented in Python.
Trainings were performed on the \server\ machine.

\section{Results}
\pgraph{Comparison with other Trackers.}
\rev{Table \ref{tab:accuracy} reports the performance of ADATRAS in comparison with teachers and other DRTs on the considered robotic domains.} 
After adaptation, our proposed tracker competes with the top-performing trackers generally. On some domains (e.g.\ptbtir, KITTI) it also outperforms them. Regarding the processing speed, ADATRAS is always faster than these methods.
The performance of TRAS \cite{Dunnhofer2020accv}, RE3 \cite{RE3} and GOTURN \cite{GOTURN} demonstrate the difficulties of DRTs due to the domain shift. In Figure \ref{fig:qualitative}, qualitative results of ADATRAS in comparison with the non adapted TRAS are presented. 
\rev{Given these results, our methodology allows to make DRTs accurate as state-of-the-art visual trackers in challenging robotic vision domains.}

We analyse in depth our methodology on the \visdrone, \ptbtir, \aquabox\ domains from now on.
Figure \ref{fig:evaltemp} shows how the performance of ADATRAS and the teachers change considering weak supervision also for $\targetsettest$. In particular, the $\ss$ and $\ps$ performance was analyzed with ground-truths $\bboxgt_t$ available every 4-th, 8-th, 16-th, 32-th, and 64-th frame. Overall, the performance tends to increase as fewer references are used for evaluation, especially for domains with a smaller number of frames. For some less-frequent $\bboxgt_t$ settings, ADATRAS results more accurate than the teachers.

\begin{table}[t]
\setlength\tabcolsep{3.3pt}
\fontsize{6}{7}\selectfont
	\centering
	\caption{Speed performance in FPS of ADATRAS and the teachers on different machines. Best results, per machine, are highlighted in bold. (ATOM w/o GPU results were not obtained because the implementation was not designed to run without it.)}
	\label{tab:speed}
	\begin{tabular}{l l | c c c | c c c | c c c}
		\toprule
		\multirow{2}{*}{Tracker} & & \multicolumn{3}{c|}{\visdrone} & \multicolumn{3}{c|}{\ptbtir} & \multicolumn{3}{c}{\aquabox} \\
            &        & \server & \desktop & \embedded & \server & \desktop & \embedded & \server & \desktop & \embedded \\
		\midrule
		
		\multirow{2}{*}{MDNet \cite{MDNet}}  & w GPU & 2 & 3 & $< 1$ & 2 & 3 & $< 1$ & 2 & 3 & $< 1$ \\
		& w/o GPU & $< 1$ & $< 1$ & $< 1$ & $< 1$ & $< 1$ & $< 1$ & $< 1$ & $< 1$ & $< 1$ \\
		\midrule
		
		\multirow{2}{*}{SiamRPN++ \cite{SiamRPNpp}}  & w GPU & 31 & 28 & 1  & 55 & 43 & 1  & 41 & 35 & 1  \\
		& w/o GPU & 3 & 3 & $< 1$  & 3 & 3 & $< 1$  & 3 & 3 & $< 1$  \\
		\midrule
		
		\multirow{2}{*}{ATOM \cite{ATOM}}  & w GPU & 16 & 28 & 3 & 24 & 59 & 4 & 20 & 43 & 4 \\
		& w/o GPU & - & - & - & - & - & - & - & - & - \\
		\midrule
		
		\multirow{2}{*}{ADATRAS}  & w GPU & \textbf{67} & \textbf{80} & \textbf{15} & \textbf{170} & \textbf{216} & \textbf{23} & \textbf{165} & \textbf{185} & \textbf{26} \\
		& w/o GPU & \textbf{33} & \textbf{55} & \textbf{2} & \textbf{62} & \textbf{91} & \textbf{2} & \textbf{98} & \textbf{168} & \textbf{2} \\

		\bottomrule		
\end{tabular}
\end{table}

\begin{table}[t]
\fontsize{7}{8}\selectfont
	\centering
	\caption{ Comparison of the proposed weakly-supervised domain adaptation method (last row) with baselines that: do not adapt; are trained from scratch; do fine-tuning. Best results, per method, are highlighted in bold.}
	\label{tab:baselines}
	\setlength\tabcolsep{.18cm}
	\begin{tabular}{l | c c | c c | c c }
		\toprule
		\multirow{2}{*}{Method} & \multicolumn{2}{c|}{\visdrone} & \multicolumn{2}{c|}{\ptbtir} & \multicolumn{2}{c}{\aquabox} \\
                    & $\ss$ & $\ps$ & $\ss$ & $\ps$ & $\ss$ & $\ps$ \\
		\midrule
		no adaptation (TRAS) & 0.384 & 0.653 & 0.432 & 0.603 & 0.522 & 0.619 \\
		from scratch by \cite{Dunnhofer2020accv} & 0.459 & 0.712 & 0.585 & 0.751 & 0.601 & 0.670 \\
		fine-tuning by \cite{Dunnhofer2020accv} & 0.549 & 0.795 & 0.659 & 0.848 & 0.569 & \textbf{0.762} \\
		from scratch by proposed & 0.475 & 0.722 & 0.625 & 0.819 & 0.549 & 0.695 \\
		adaptation by proposed & \textbf{0.552} & \textbf{0.823} & \textbf{0.661} & \textbf{0.862} & \textbf{0.576} & 0.732 \\
		\bottomrule		
\end{tabular}
\end{table}

\pgraph{Speed Analysis.}
Table \ref{tab:speed} reports the analysis on the processing speed of our method in comparison with the teachers on different machines. ADATRAS results the fastest method on all machine setups. Very high speeds are reached on top machines with a GPU (\server\ or \desktop), leaving large space for real-time downstream application development \rev{and making it good for lower resource robots}.
Remarkably, ADATRAS achieves real-time speed even without a GPU when run on top machines. When run on small embedded devices like \embedded, ADATRAS achieves real-time speeds on \ptbtir\ and \aquabox\, and a quasi-real-time speed on \visdrone, considering 20 FPS as real-time baseline \cite{VOT2019,VOT2020}.

\begin{figure}[t]%
\centering
\includegraphics[width=\columnwidth]{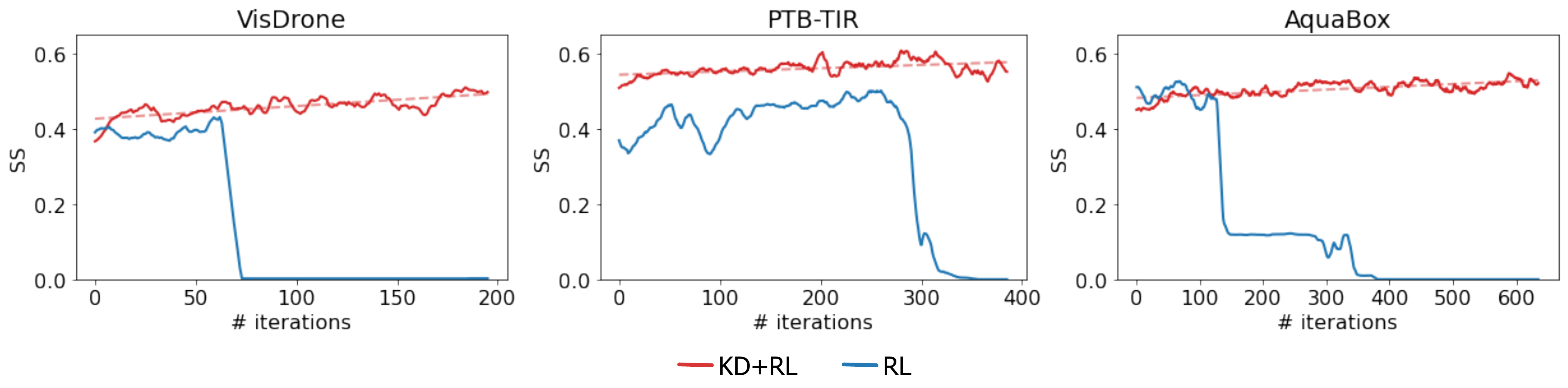}
\caption{$\ss$ trends on $\targetsettest$ of the proposed adaptation strategy (red line) and a pure RL fine-tuning (blue line) at different iterations during the learning phase. After some iterations, the RL-based solution diverges, while the proposed adaptation continuously improves ADATRAS (as shown by the red dashed line). Lines have been smoothed for better visualization.}
\label{fig:kdrl}
\end{figure}

\begin{table}[t]
\fontsize{7}{8}\selectfont
	\centering
	\caption{\rev{Performance comparison between ADATRAS and trackers fine-tuned on $\targetsettrain$. Best results, per tracker, are highlighted in bold.}}
	\label{tab:finetuning}
	\setlength\tabcolsep{.18cm}
	\begin{tabular}{l | c c | c c | c c }
		\toprule
		\multirow{2}{*}{Tracker} & \multicolumn{2}{c|}{\visdrone} & \multicolumn{2}{c|}{\ptbtir} & \multicolumn{2}{c}{\aquabox} \\
                    & $\ss$ & $\ps$ & $\ss$ & $\ps$ & $\ss$ & $\ps$ \\
		\midrule
		SiamRPN++-ft \cite{SiamRPNpp} & \textbf{0.594} & \textbf{0.842} & 0.572 & 0.733 & 0.559 & 0.703 \\
		GOTURN-ft \cite{GOTURN} & 0.380 & 0.642 & 0.353 & 0.548 & 0.450 & 0.374 \\
		RE3-ft \cite{RE3} & 0.168 & 0.297 & 0.408 & 0.554 & 0.530 & 0.586 \\
		ADATRAS & 0.552 & 0.823 & \textbf{0.661} & \textbf{0.862} & \textbf{0.576} & \textbf{0.732} \\
		\bottomrule		
\end{tabular}
\end{table}

\pgraph{Comparison with Baselines.}
Table \ref{tab:baselines} presents the performance of our methodology in comparison with baseline adaptation and no-adaptation methods. Our method (in the last row) outperforms the tracker without adaptation (TRAS \cite{Dunnhofer2020accv}) on every target domain. 
These results show that the goal of improving the baseline tracker's accuracy with weak supervision is achieved. 
Generally, performance is even improved with respect to the dense bounding-box supervision experiments performed following \cite{Dunnhofer2020accv}, thus justifying the introduced improvements. 
Adapting past knowledge is effective to reduce overfitting, as demonstrated by the improvement over the results reported in rows two and four, for which training was performed from scratch.
Figure \ref{fig:kdrl} shows how the adaptation procedure with only an RL signal causes the student to diverge after some iterations. This is probably due to the increased length of the interaction (based on the curriculum strategy) that causes wrong gradient estimations.
\rev{Table \ref{tab:finetuning} shows the performance of ADATRAS in comparison with other trackers fine-tuned (following their original learning strategy) on $\targetsettrain$. 
Our approach results generally better. This can be attributed to the RL strategy, which leads to more efficient data exploration, ultimately providing a data augmentation effect.}

\begin{table}[t]
\fontsize{7}{8}\selectfont
	\centering
	\caption{Performance of the proposed tracker under different supervision settings. Best results, per supervision method, are highlighted in bold.}
	\label{tab:weaksup}
	\begin{tabular}{l | c c | c c | c c }
		\toprule
		\multirow{2}{*}{Supervision} & \multicolumn{2}{c|}{\visdrone} & \multicolumn{2}{c|}{\ptbtir} & \multicolumn{2}{c}{\aquabox} \\
                    & $\ss$ & $\ps$ & $\ss$ & $\ps$ & $\ss$ & $\ps$ \\
		\midrule
		GT $\bboxgt_t$ & 0.490 & 0.757 & 0.640 & 0.803 & 0.517 & 0.629 \\
		KD $\bbox_t^{(\teacher)}$ & 0.497 & 0.769 & 0.601 & 0.788 & 0.514 & 0.634 \\
		$\weaksupiou(\cdot)$ & \textbf{0.552} & 0.823 & \textbf{0.661} & 0.862 & \textbf{0.576} & 0.732 \\
		$\weaksupdist(\cdot)$ & 0.523 & \textbf{0.852} & 0.638 & \textbf{0.894} & 0.575 &\textbf{ 0.774} \\
		
		\bottomrule		
\end{tabular}
\end{table}

\begin{figure}[t]%
\centering
\includegraphics[width=\columnwidth]{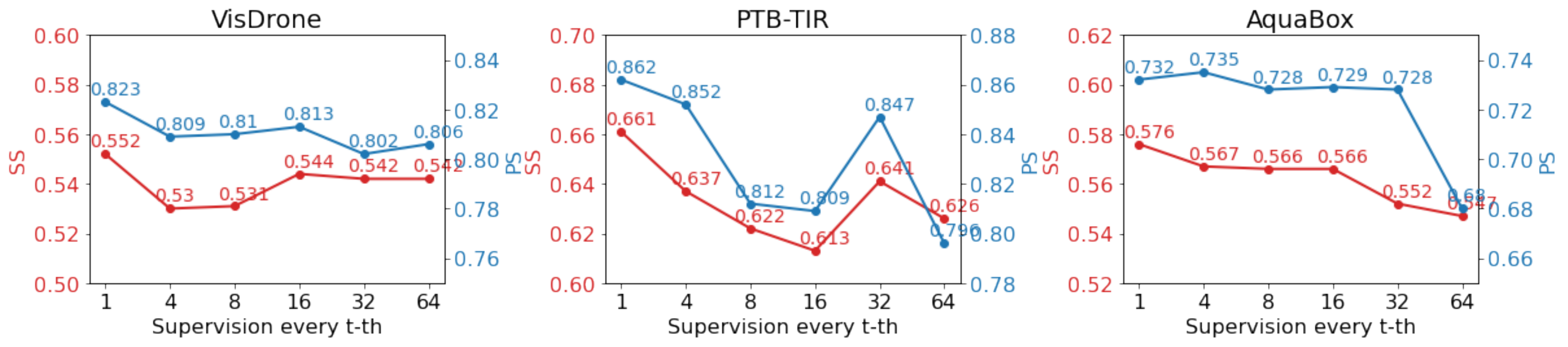}
\caption{Performance of ADATRAS trained on the three domains considering weak supervision delayed at different steps in time.}
\label{fig:weaksuptemp}
\end{figure}

\pgraph{Weak Supervision Analysis.}
Table \ref{tab:weaksup} reports the performances of ADATRAS adapted with different kind of supervision. The functions $\weaksupiou(\cdot), \weaksupdist(\cdot)$ improve by a good margin the results achieved by learning from ground-truth bounding-boxes (GT $\bboxgt_t$), or by learning just from the bounding-box predictions given by the teachers (KD $\bbox_t^{(\teacher)}$). Using $\weaksupiou(\cdot)$ allows to achieve the best results in $\ss$, while using $\weaksupdist(\cdot)$ improves the $\ps$ performance. These results confirm that optimizing a specific performance measure as reward function induces the improvement of such measure at test time. 
Moreover, we analyzed the sensibility of ADATRAS to the weak supervision delayed in time. In particular, ADATRAS was trained with weak supervision happening every 4-th, 8-th, 16-th, 32-th, 64-th, temporal step $t$ (in a 20 FPS setting, this would mean every 0.2, 0.4, 0.8, 1.6, 3.2, seconds). Results are shown in Figure \ref{fig:weaksuptemp}. In general, performance tends to decrease as the supervision is more delayed, but without a significant loss, especially for domains with a larger number of frames (e.g. \visdrone). 
Interestingly, particularly delayed supervision allows achieving similar performance as the case in which supervision is given more frequently. 
We hypothesize this is due to the distribution between supervised $\targetsettrain$'s frames and the frames appearing in $\targetsettest$.
More importantly, our proposed adaptation strategy reaches and surpasses the GT $\bboxgt_t$ adaptation (row one of Table \ref{tab:baselines}) even with delayed supervision.

\pgraph{\rev{Impact of Source and Teachers.}}
Table \ref{tab:source} reports the performance of ADATRAS which $\student(\cdot|\weights)$ is adapted after learning on two different source domains. 
The performance of the two settings before adaptation are reported in the first two rows. $\student(\cdot|\weights)$ trained on the GOT-10k dataset performs much better than $\student(\cdot|\weights)$ optimized on the LaSOT dataset, due to the broader knowledge acquired on the larger GOT-10k. The results in rows three and four show that such a trend is maintained also after adaptation, showing that a better baseline tracking behavior is an important factor to achieve a better adapted tracking policy.

\begin{table}[t]
\fontsize{7}{8}\selectfont
	\centering
	\caption{Performance comparison of ADATRAS adapted starting from different source domain knowledge. Performance without adaptation is also reported in the first two rows. Best results are highlighted in bold.}
	\label{tab:source}
	\setlength\tabcolsep{.18cm}
	\begin{tabular}{l | c c | c c | c c }
		\toprule
		\multirow{2}{*}{Source} & \multicolumn{2}{c|}{\visdrone} & \multicolumn{2}{c|}{\ptbtir} & \multicolumn{2}{c}{\aquabox} \\
                    & $\ss$ & $\ps$ & $\ss$ & $\ps$ & $\ss$ & $\ps$ \\
		\midrule
		no-adaptation - LaSOT & 0.233 & 0.475 & 0.359 & 0.591 & 0.328 & 0.335 \\
		no-adaptation - GOT-10k & 0.384 & 0.653 & 0.432 & 0.603 & 0.522 & 0.619 \\
		adaptation - LaSOT  & 0.475 & 0.726 & 0.646 & 0.860 & 0.541 & 0.719 \\
		adaptation - GOT-10k & \textbf{0.552} & \textbf{0.823} & \textbf{0.661} & \textbf{0.862} & \textbf{0.576} & \textbf{0.732} \\
		\bottomrule		
\end{tabular}
\end{table}

\begin{table}[t]
\fontsize{7}{8}\selectfont
	\centering
	\caption{Performance of the proposed tracker with different teacher setups. Best results, per setup, are highlighted in bold.}
	\label{tab:teachers}
	\setlength\tabcolsep{.2cm}
	\begin{tabular}{l | c c | c c | c c }
		\toprule
		\multirow{2}{*}{Teachers} & \multicolumn{2}{c|}{\visdrone} & \multicolumn{2}{c|}{\ptbtir} & \multicolumn{2}{c}{\aquabox} \\
                    & $\ss$ & $\ps$ & $\ss$ & $\ps$ & $\ss$ & $\ps$ \\
		\midrule
		$\tm$ & 0.525 & 0.786 & 0.542 & 0.733 & 0.521 & 0.556 \\
		$\ts$ & 0.467 & 0.731 & 0.576 & 0.774 & 0.582 & 0.642 \\
		$\ta$ & 0.466 & 0.734 & 0.604 & 0.755 & 0.537 & 0.729 \\
		$\tp$ random selection & 0.533 & 0.791 & 0.654 & 0.825 & \textbf{0.603} & \textbf{0.735} \\
		$\tp$ with Eq. (\ref{eq:teacherselection}) & \textbf{0.552} & \textbf{0.823} & \textbf{0.661} & \textbf{0.862} & 0.576 & 0.732 \\
		
		\bottomrule		
\end{tabular}
\end{table}
In Table \ref{tab:teachers} ADATRAS was analyzed after adaptation with different teacher setups. Using a better teacher does not translate into better performance, suggesting that it is important to understand on which sequences teachers perform well. Best results are obtained by learning from multiple teachers, demonstrating that the knowledge of these is compensated on $\targetsettrain$.
Moreover, giving a ranking of teachers and selecting the best for each training video, allows better performance on domains where teachers are better and more training frames are available (e.g. \visdrone\ and \ptbtir). 
For some domain, selecting them randomly leads to almost equal performances than the ranking-based selection, probably due to the data distributions of $\targetsettrain$ and $\targetsettest$ which don't reflect the average performance of the teachers.

\section{Conclusions}
In this paper, we present the first methodology for domain adaption of DRTs, which are fast but inaccurate methods. We achieve such a goal by proposing a weakly-supervised approach, thus reducing labeling effort. RL is used to express weak supervision as a scalar application-dependent and temporally-delayed feedback.
KD is employed to achieve convergence, and to transfer knowledge from other trackers. 
Extensive experiments on five different domains and three machine setups demonstrate the effective usage for various robotic perception domains. Real-time speed is achieved on small embedded devices and on machines without GPUs. Accuracy is comparable to more powerful but slow state-of-the-art trackers.

\ifCLASSOPTIONcaptionsoff
  \newpage
\fi

\bibliographystyle{IEEEtran}
\bibliography{IEEEabrv,egbib}

\end{document}